\documentclass[acmtog]{acmart}

\setcitestyle{square} 

\usepackage{graphicx}
\usepackage{comment}
\usepackage{amsmath,amssymb} 
\usepackage{color}
\usepackage{multirow}
\long\def\ignorethis#1{}

\usepackage{color}
\definecolor{gray}{rgb}{0.35,0.35,0.35}
\definecolor{blue}{rgb}{0,0,1}
\definecolor{white}{rgb}{1,1,1}
\definecolor{black}{rgb}{0,0,0}
\definecolor{dark-brown}{rgb}{0.2,0.1,0}
\definecolor{orange}{rgb}{1.0,0.26,0}

\newcommand{\whitetxt}[1]{{\color{white}#1}\normalfont}

\newbox\jsavebox
\newcommand{\jsubfig}[2]{%
	\sbox\jsavebox{#1}%
	\parbox[t]{\wd\jsavebox}{\centering\usebox\jsavebox\\#2}%
	}

\usepackage[ruled]{algorithm2e} 

\SetAlFnt{\small}
\SetAlCapFnt{\small}
\SetAlCapNameFnt{\small}
\SetAlCapHSkip{0pt}

\begin{document}

\title{Co-occurrence Based Texture Synthesis} 

%
\author{Anna Darzi, Itai Lang, Ashutosh Taklikar}
\affiliation{%
Tel Aviv University}%
\author{Hadar Averbuch-Elor}
\affiliation{%
Cornell Tech, Cornell University}%
\author{Shai Avidan}
\affiliation{%
Tel Aviv University}%
%

\begin{abstract}
As image generation techniques mature, there is a growing interest in explainable representations that are easy to understand and intuitive to manipulate.
In this work, we turn to co-occurrence statistics, which have long been used for texture analysis, to learn a controllable texture synthesis model.
We propose a fully convolutional generative adversarial network, conditioned locally on co-occurrence statistics, to generate arbitrarily large images while having local, interpretable control over the texture appearance.
To encourage fidelity to the input condition, we introduce a novel differentiable co-occurrence loss that is integrated seamlessly into our framework in an end-to-end fashion.
We demonstrate that our solution offers a stable, intuitive and interpretable latent representation for texture synthesis, which can be used to generate a smooth texture morph between different textures.
We further show an interactive texture tool that allows a user to adjust local characteristics of the synthesized texture image using the co-occurrence values directly.

\ignorethis{We model local texture patterns using the co-occurrence statistics of pixel values. We then train a generative adversarial network, conditioned on co-occurrence statistics, to synthesize new textures from the co-occurrence statistics and a random noise seed. Co-occurrences have long been used to measure similarity between textures. That is, two textures are considered similar if their corresponding co-occurrence matrices are similar. By the same token, we show that multiple textures generated from the same co-occurrence matrix are similar to each other. This gives rise to a new texture synthesis algorithm. 

We show that co-occurrences offer a stable, intuitive and interpretable latent representation for texture synthesis. Our technique can be used to generate a smooth texture morph between two textures, by interpolating between their corresponding co-occurrence matrices. We further show an interactive texture tool that allows a user to adjust local characteristics of the synthesized texture image using the co-occurrence values directly.}

\end{abstract}


\maketitle

\section{Introduction}

Deep learning has revolutionized our ability to generate novel images. 
Most notably, generative adversarial networks (GANs) have shown impressive results in various domains, including textures. Nowadays, GANs can generate a distribution of texture images that is often indistinguishable from the distribution of real ones.
The goal of the generator is conceptually simple and boils down to training the network weights 
to map a latent code, sampled from a random distribution, 
to realistic samples.

However, generative networks are typically non-explainable and hard to control. That is, given a generated sample, it is generally hard to explain its latent representation and to directly modify it to output a new sample with desired properties.

%
%

\ignorethis{As modern solutions commonly struggle with providing the user with such fine-grained control,} Recently, we are witnessing growing interests in interpreting the latent space of GANs, aspiring to better understand its behaviour \cite{radford2015unsupervised,bau2019visualizing,shen2019interpreting}.
These studies tap into vector arithmetics in latent space, or use more complicated entangled properties to achieve better control of the texture generation process. However, many questions remain open, and local control is still difficult to achieve. 


\begin{figure}
\begin{center}
    \includegraphics[width=1.0\columnwidth]{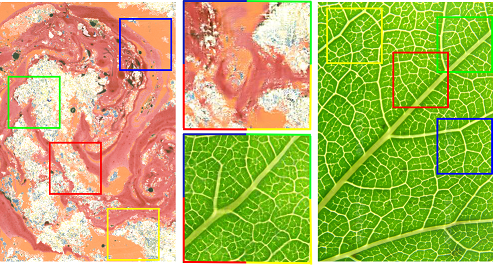}
    \end{center}
    \caption{Given a texture exemplar (left and right), co-occurrence statistics are collected for different texture crops. Using these co-occurrences, our method can synthesize a variety of novel textures with desired local properties (center), similar to those of the corresponding crops from the texture exemplar. 
    Above we demonstrate textures synthesized from co-occurrence statistics collected at four random crops (marked in unique colors). 
    }
    \label{fig:teaser}
\end{figure}

In this work, we seek a generative model for textures that is intuitive to understand and easy to edit and manipulate. 
Our key insight is that we can bypass the need to understand a highly entangled latent space by using a structured latent space, with meaningful interpretable vectors.

We use a statistical tool, co-occurrences, to serve as an encoding of texture patches.
Co-occurrences, first introduced by Julesz~\cite{Julesz:62}, capture the local joint probability of pairs of pixel values to co-occur together. They have long been used to analyze textures \cite{HaralickSD73,IsolaZKA14}. In our work, we take the opposite direction. Given a co-occurrence matrix, we can generate a variety of texture images that match it.

Technically, we train a fully convolutional conditional GAN (cGAN), conditioned on local co-occurrence statistics. 
The convolutional architecture allows for arbitrarily large generated textures, while local co-occurrence statistics gives us explainable control over \textit{local} texture appearance.
This allows synthesizing a variety of inhomogeneous textures, such as those shown in Figure \ref{fig:teaser}, by conditioning locally on different co-occurrences collected from the input textures. 

To enable end-to-end training using co-occurrence statistics, we introduce a new differentiable \emph{co-occurrence loss}, which penalizes inconsistencies between the co-occurrence of the synthesized texture and the input condition. 
We demonstrate that our proposed training objective allows for a simple and interpretable latent representation, without compromising fidelity on a large variety of inhomogeneous textures. 
We show that our approach can be used, for example, to interpolate between two texture regions by interpolating between their corresponding co-occurrence matrices, resulting in a {\em dynamic} texture morph. We also illustrate how to control the generated texture by editing the co-occurrence input.

\textbf{Our contributions} are twofold: First, we introduce a stable, intuitive and interpretable latent representation for texture synthesis, offering parametric control over the synthesized output using co-occurrence statistics. Second, we present a fully convolutional cGAN architecture with a differentiable co-occurrence loss, enabling end-to-end training. Our code is publicly available at \url{https://github.com/coocgan/cooc_texture/}.


\section{Related Work}
There are different ways to model texture. Heeger and Bergen \cite{Heeger1995} represented textures as histograms of different levels of a steerable pyramid, while De Bonet~\cite{Bonet97} represented texture as the conditional probability of pixel values at multiple scales of the image pyramid. New texture is synthesized by matching the statistics of a noise image to that of the input texture. Portilla and Simoncelli~\cite{portilla2000parametric} continued this line of research, using a set of statistical texture properties. None of these methods used co-occurrence statistics for texture synthesis.

Stitching-based methods \cite{efros2001image,kwatra2003graphcut,kwatra2005texture} assume a non-parametric model of texture. In this case a new texture is generated by sampling patches from the input texture image. This sampling procedure can lead to deteriorated results and one way to fix that is to use an objective function that forces the distribution of patches in the output texture to match that of the input texture \cite{wexler2007space,SimakovCSI08}. These methods were proved to be very effective in synthesizing plausible textures. 

Some methods to interpolate between textures and not necessarily synthesize new texture have been proposed. For example, Matusik \etal~\cite{Matusik:2005:TDU} capture the structure of the induced space by a simplicial complex where vertices of the simplices represent input textures. Interpolating between vertices corresponds to interpolating between textures. Rabin \etal~\cite{Rabin2010} interpolate between textures by averaging discrete probability distributions that are treated as a barycenter over the Wasserstein space. Soheil \etal~\cite{Darabi12:ImageMelding12} use the screened Poisson equation solver to meld images together and, among other applications, show how to interpolate between different textures.

Deep learning for texture synthesis by Gatys \etal~\cite{gatys2015texture} follows the approach of Heeger and Bergen~\cite{Heeger1995}. Instead of matching the histograms the image pyramid, they match the Gram matrix of different features maps of the texture image, where the Gram matrix measures the correlation between features at selected layers of a neural network. This approach was later improved by subsequent works \cite{ulyanov2016texture,sendik2017deep}. These methods look at the pair-wise relationships between features, which is similar to what we do. However, the Gram matrix measures the correlation of deep features, whereas we use the co-occurrence statistics of pixel values.

Alternatively, one can use a generative adversarial network (GAN) to synthesize textures that resemble the input exemplar. Li and Wand~\cite{li2016precomputed} used a GAN combined with a Markov Random Field to synthesize texture images from neural patches. Liu \etal~\cite{liu2016texture} improved the method of Gatys \etal~\cite{gatys2015texture} by adding constraints on the Fourier spectrum of the synthesized image, and Li \etal~\cite{Li2017:DiversifiedTexture} use a feed-forward network to synthesize diversified texture images. Zhou \etal~\cite{zhou2018non} use GANs to spatially expand texture exemplars, extending non-stationary structures. Fr{\"u}hst{\"u}ck \etal~\cite{fruhstuck2019tilegan} synthesize large textures using a pre-trained generator, which can produce images at higher resolutions.

Texture synthesis using GANs is also used for texture interpolation. Jetchev \etal~\cite{jetchev2016texture} suggested using a spatial GAN (SGAN), where the input noise to the generator is a spatial tensor rather than a vector. This method was later extended to the periodic spatial GAN (PSGAN), proposed by Bergmann \etal~\cite{bergmann2017learning}. Interpolating between latent vectors within the latent tensor results in a spatial interpolation in the generated texture image. These works focus on spatial interpolation, and learn the input texture's structure as a whole. We focus on representing the appearance of different local regions of the texture in a controllable manner.

The most related work to ours is the recent Texture Mixer of Yu \etal~\cite{yu2019texture}, which aims at texture interpolation in the latent space. However, different from our work, their method requires encoding sample crops to the latent space and control is obtained in the form of interpolating between the representations of these sample crops. Our method, on the other hand, provides a parametric and interpretable latent representation which can be controlled directly.


Co-occurrences were introduced by Julesz~\cite{Julesz:62} who conjectured that two distinct textures can be spontaneously discriminated based on their second order statistics (i.e., co-occurrences). This conjecture is not necessarily true because it is possible to construct distinct images that have the same first (histograms) and second order (co-occurrence) statistics. It was later shown by Yellott~\cite{Yellott:93} that images with the same third-order statistics are indistinguishable. In practice, co-occurrences have long been used for texture analysis \cite{HaralickSD73,IsolaZKA14}, but not for texture synthesis, as we propose in this work.

\section{Method} \label{sec:method}

\begin{figure}
\begin{center}
\centering
\includegraphics[width=0.9\linewidth]{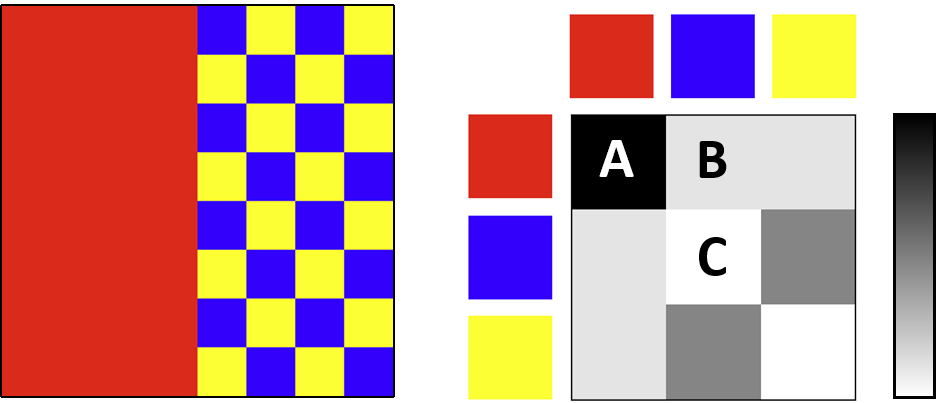}
\end{center}
\caption{{\bfseries Co-occurrence Example.} An input image (left) and its corresponding co-occurrence matrix (right), computed according to Equations~\ref{eq:cooc_collection} and~\ref{eq:soft_assignment}. In this example we assume that co-occurrence is only measured using a 4-neighborhood connectivity. Red pixels appear very frequently next to each other, so their co-occurrence measure is high (bin A). The red pixels appear less frequently with blue ones, which is reflected as a lower value in the co-occurrence matrix (bin B). Blue pixels do not appear next to each other, yielding a zero probability (bin C).}  
\label{fig:cooc_matrix_example}
\end{figure}

\begin{figure*}
\begin{center}
\includegraphics[width=0.85\linewidth]{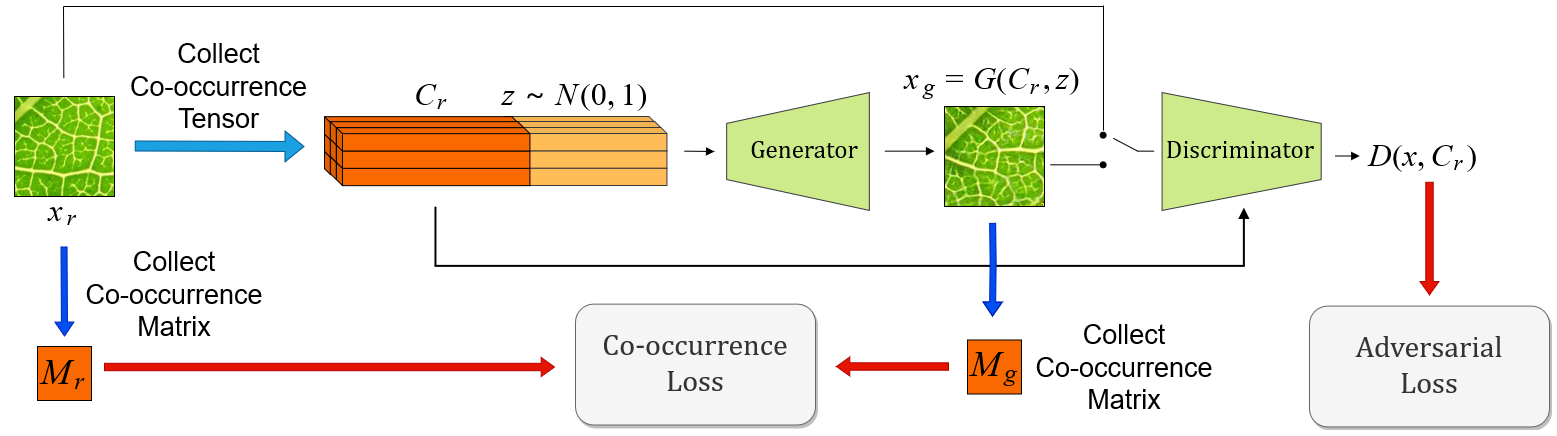}
\end{center}
\caption{{\bfseries Method Overview.} The generator receives as input a co-occurrence tensor $C_r$, corresponding to a real texture crop $x_r$. This tensor is concatenated to a random noise tensor $z$. The generator outputs a synthetic texture sample $x_g$. The discriminator's input alternates between real and synthetic texture samples. It also receives the co-occurrence tensor $C_r$. In addition, the co-occurrence of the synthesized texture is required to be consistent with the input statistics to the generator. By this architecture, the generator learns to synthesize samples with desired local texture properties, as captured by the co-occurrence tensor.}

\label{fig:cGAN_overview}
\end{figure*}

We use a conditional generative adversarial network (cGAN) to synthesize textures with spatially varying co-occurrence statistics. Before diving into the details, let us fix notations first. A texture {\em patch} (i.e., a $64 \times 64$ pixels region) is represented by a local co-occurrence matrix. A texture {\em crop} (i.e., a $128 \times 128$ pixels region) is represented by a collection of local co-occurrence matrices, organized as a co-occurrence tensor. 
We train our cGAN on texture crops, because crops capture the interaction of neighboring co-occurrences. This allows the generator to learn how to synthesize texture that fits spatially varying co-occurrence statistics. Once the cGAN is trained, we can feed the generator with a co-occurrence tensor and a random seed to synthesize images of arbitrary size. We can do that because both the discriminator and generator are fully convolutional.

An overview of our cGAN architecture is shown in Figure~\ref{fig:cGAN_overview}. It is based on the one proposed by Bergmann \etal~\cite{bergmann2017learning}. In what follows, we explain how the co-occurrence statistics are collected, followed by a description of our cGAN and how we use it to generate new texture images.

We use the co-occurrence matrix to represent the local appearance of patches in the texture image. For a given patch, we calculate the co-occurrence matrix $M$ of the joint appearance statistics of pixel values \cite{jevnisek2017cooc,kat2018matching}.

The size of the co-occurrence matrix scales quadratically with the number of pixel values and it is therefore not feasible to work directly with RGB values. To circumvent that we quantize the color space to a small number of clusters. The pixel values of the input texture image are first grouped into $k$ clusters using standard $k$-means. This results in a $k \times k$ co-occurrence matrix. 

Let $(\tau_a, \tau_b)$ denote two cluster centers. Then $M(\tau_a, \tau_b)$ is given by:

\begin{equation}\label{eq:cooc_collection}
M(\tau_a, \tau_b)=\frac{1}{Z}\sum_{p,q}exp(-\frac{d(p,q)^2}{2\sigma^2})K(I_p, \tau_a)K(I_q, \tau_b) ,
\end{equation}

\noindent where $I_p$ is the pixel value at location $p$, $d(p,q)$ is the Euclidean distance between pixel locations $p$ and $q$, $\sigma$ is a user specified parameter, and $Z$ is a normalizing factor designed to ensure that the elements of $M$ sum up to $1$. 

$K$ is a soft assignment kernel function that decays exponentially with the distance of the pixel value from the cluster center:

\begin{equation}\label{eq:soft_assignment}
K(I_p, \tau_l)=exp(-\sum_i\frac{(I_p^i - \tau_l^i)^2}{(\sigma_l^i)^2}) ,
\end{equation}

\noindent where $i$ runs over the RGB color channels and $\sigma_l^i$ is the standard deviation of color channel $i$ of cluster $l$.

The contribution of a pixel value pair to the co-occurrence statistics decays with their Euclidean distance in the images plane. In practice, we do not sum over all pixel pairs $(p,q)$ in the image, but rather consider only pixels $q$ within a window around $p$. An illustrative image and its corresponding co-occurrence matrix is given in Figure~\ref{fig:cooc_matrix_example}.

We collect the co-occurrence statistics of an image crop $x$ of spatial dimensions $h\times w$, according to Equations~\ref{eq:cooc_collection} and~\ref{eq:soft_assignment}. We row-stack each of these matrices to construct a co-occurrence volume of size $h\times w \times k^2$. This volume is then downsampled spatially by a factor of $s$, to obtain a co-occurrence tensor $C$. This downsampling is meant to allow more variability for the texture synthesis process, implicitly making every spatial position in $C$ a description of the co-occurrence of a certain receptive field in $x$.

\subsection{Texture Synthesis}
We denote a real texture crop as $x_r$, and its co-occurrence tensor $C_r$. The random noise seed tensor is denoted as $z$, where its entries are drawn from a normal distribution $\mathcal{N}(0,1)$. In order to balance the influence of the co-occurrence and the noise on the output of the generator, we also normalise $C_r$ to have zero mean and unit variance. 

The concatenated co-occurrence and noise tensor are given as input to the generator. The output of the generator $G$ is a synthesized crop marked as $x_g = G(C_r, z)$. The corresponding downsampled co-occurrence of $x_g$ is marked $C_g$.

The input to the discriminator $D$ is a texture crop that is either a real crop from the texture image ($x_r$) or a synthetic one from the generator ($x_g$). In addition to the input texture crop, the original co-occurrence tensor $C_r$ is also provided to the discriminator. In case of a real input texture, this is its corresponding co-occurrence tensor. For a synthetic input, the same co-occurrence condition given to the generator is used at the discriminator.

We emphasize that both the generator and discriminator are conditioned on the co-occurrence tensor. With this cGAN architecture, the discriminator teaches the generator to generate texture samples that preserve local texture properties, as captured by the co-occurrence tensor. In order to further guide the generator to respect the co-occurrence condition when synthesizing texture, we compute the co-occurrence statistics of the generated texture and demand consistency with the corresponding generator's input. 

We train the network with Wasserstein GAN with gradient penalty (WGAN-GP) objective function, proposed by Gulrajani \etal~\cite{gulrajani2017improved}. In addition, we add a novel consistency loss on the co-occurrence of the generated texture and the input condition. 

The generator is optimized according the the following loss function:
\begin{equation}\label{eq:g_objective}
L(G) = -\mathbb{E}_{x_g \sim \mathbb{P}_g} [D(G(z,C_r),C_r)] + \lambda_M |M_g - M_r|_1 ,
\end{equation}

\noindent where $D$ is the discriminator network, and $|\cdot|_1$ is $L_1$ loss. 

The discriminator is subject to the following objective:
\begin{equation}\label{eq:cd_objective}
\begin{split}
L(D) = \mathbb{E}_{x_r \sim \mathbb{P}_r} [D(x_r,C_r)] - \mathbb{E}_{x_g \sim \mathbb{P}_g} [D(x_g,C_r)]& + \\
\lambda_p \mathbb{E}_{\hat{x} \sim \mathbb{P}_{\hat{x}}} [(||\nabla_{\hat{x}} D(\hat{x},C_r)||_2 - 1)^2]& ,
\end{split}
\end{equation}

\noindent where $\hat{x}$ is a randomly sampled as a liner interpolation between real and synthetic textures $\hat{x}=u x_r + (1-u) x_g$, with a uniform interpolation coefficient $u \sim U[0,1]$.

The training is done with batches of texture samples and their corresponding co-occurrence tensors. When training the discriminator, the co-occurrences of the real texture samples are used to generate the synthetic samples. For the generator training, only the co-occurrences of the batch are used.


\begin{figure}
\begin{center}
\rotatebox[origin=l]{90}{\whitetxt{ssssssssssssssS}Random seed interpolation}
\includegraphics[width=0.85\linewidth]{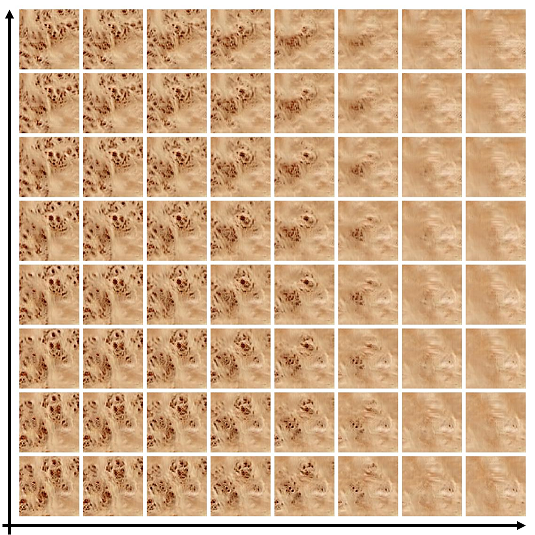} 
\vspace{-5pt} \\
Co-occurrence interpolation
\end{center}
\caption{Our technique synthesizes textures given a co-occurrence matrix and a random seed vector. Above we illustrate textures generated from interpolating between two co-occurrence vectors (across the columns) and two random seed vectors (across the rows).
}
\label{fig:interp_z_c}
\end{figure}

\begin{figure}
\begin{center}
	\includegraphics[width=1.0\columnwidth]{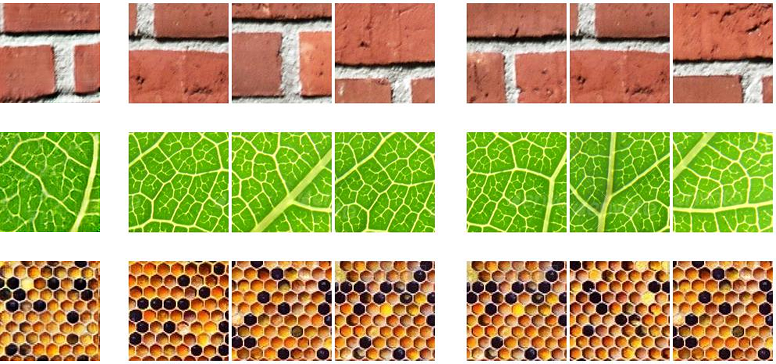} \\
Output \whitetxt{ssssssssss} NNs (RGB) \whitetxt{ssssssssdwsss} NNs (co-occurrence) \whitetxt{sss}
\end{center}
\caption{{\bfseries Novelty of generated samples.} Left: Generated textures using co-occurrences from the test set. For each generated sample, we find its nearest neighbors (NN) in the training set, measured in terms of RGB values (middle) and co-occurrences (right). As illustrated above, the generated textures are different from their nearest neighbors in the training set. }

\label{fig:closest_training_crops}
\end{figure}

\section{Results and Evaluation} \label{sec:evaluation}
We evaluate our technique on a wide variety of textures, including from the Describable Textures Dataset \cite{cimpoi2014describing}, from the work of  Bellini \etal~\cite{bellini2016time} and also several texture images we found online.
We demonstrate the performance of our method in terms of: (i) fidelity and diversity, (ii) novelty and (iii) stability. 
Additionally, we compare our approach to previous works (Section \ref{sec:previous_work}). 
We also perform an ablation study, demonstrating the importance of the different components in our work (Section \ref{sec:ablations}). 
Finally we conclude by discussing limitations of our presented approach (Section \ref{sec:limitations}).

\begin{figure*}
\begin{center}
\includegraphics[width=1.0\textwidth]{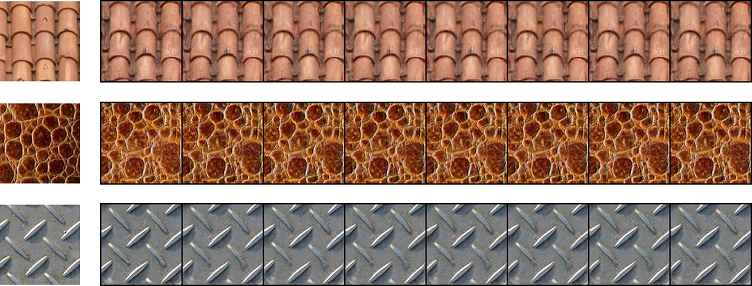}
\end{center}
\caption{{\bfseries Stability test.} Textures were generated in a loop. In the first iteration, co-occurrences are taken from the test set. In the next, co-occurrence of the generated texture is used. The original crop from the texture exemplar is on the left and the generated sequence for eight iterations is on the right. It can be noted that the textures remain stable over the iterations.}
\label{fig:cooc_loop}
\end{figure*}

\subsection{Experimental Details}
For all texture images we keep $k$ either equal to $2$, $4$ or $8$ clusters. We collect the co-occurrence statistics for each pixel. We take a patch of $65 \times 65$ around that pixel, and calculate the co-occurrence statistics using a window of size $51 \times 51$ around each pixel in that patch and set $\sigma^2$ in Equation~\ref{eq:cooc_collection} to $51$. So, for each pixel we have $k \times k$ co-occurrence matrix, which is reshaped to $k^2$, and for an image of $w \times h \times 3$, we have a co-occurrence volume of $w \times h \times k^2$.

The dataset for a texture image contains $N=2,000$ crops of $n \times n$ pixels. As our work aims at extracting and analyzing local properties, we use $n=128$. The down sampling factor of the co-occurrence volume is $s=32$. Thus, the spatial dimensions of the condition co-occurrence tensor are $4 \times 4$.

The architecture of the generator and discriminator is fully convolutional, thus larger textures can be synthesized at inference time. The architecture is based on that of PSGAN~\cite{bergmann2017learning}. The generator in our case is a stack of $5$ convolutional layers, each having an upsampling factor of $2$ and filter of size $5 \times 5$, stride of $1$ and a padding of $2$, with ReLU activation and batch normalization.

The discriminator too is a stack of $5$ convolutional layers, with filter size of $5 \times 5$, stride of $2$ and padding of $2$, with sigmoid activation. The stride here is $2$ to bring down the upscaled spatial dimensions by the generator back to original input size. After the activation of the third layer, we concatenate the co-occurrence volume in the channel dimension, to help the discriminator condition on the co-occurrence as well (see Figure~\ref{fig:cGAN_overview}).

For each texture image, we train our method for $120$ epochs using Adam optimizer with momentum of $0.5$. The learning rate is set to $0.0002$, and the gradient penalty weight is $\lambda=1$. The training time is about $3$ hours using a NVIDIA 1080Ti GPU. 

\subsection{Evaluation Tests} \label{sec:evaluation_tests}

\paragraph*{Fidelity and Diversity.}
Texture synthesis algorithms must balance the contradicting requirements of fidelity and diversity. Fidelity refers to how similar the synthesized texture is to the input texture. Diversity expresses the difference in the resulting texture, with respect to the input texture. We maintain fidelity by keeping the co-occurrences fixed and achieve diversity by varying the noise seed. 

Figure~\ref{fig:teaser} shows that the generated samples resemble their corresponding texture crop. Utilizing the adversarial loss during training can be viewed as an implicit demand of this property, while the the co-occurrence loss requires is more explicitly. Additional fidelity and diversity results are presented in the supplementary. 

We demonstrate the smoothness of the latent space by interpolating in two axes: between co-occurrence tensors and between noise tensors (see Figure \ref{fig:interp_z_c}). Interpolation between two different co-occurrences results in a smooth traversal between different texture characteristics. On the other hand, for a given co-occurrences tensor, interpolation between different noise tensors smoothly transitions between texture samples with similar properties. In addition, any intermediate result of the interpolation yields a plausible texture result. This suggests that the generator learned a smooth texture manifold, with respect to both co-occurrence and noise.

\paragraph*{Novelty of texture samples.} 
To verify that our network is truly generating novel, unseen texture samples and not simply memorizing seen examples, we examine the nearest neighbors in the training set. To do so, we generate textures using unseen co-occurrences from the test set and search for their nearest neighbors in the training set, in terms of $L_1$ of RGB values. For comparison, we also compute the co-occurrence of the generated samples and look for nearest neighbor, in $L_1$ sense, in the co-occurrence space. 

We demonstrate the results of this experiment in Figure ~\ref{fig:closest_training_crops}. The spatial arrangement of nearest neighbors in RGB space resembles that of the synthesized texture, yet they are not identical. Nearest neighbors in the co-occurrence space may have different spatial arrangement. In any case, while the training samples bear some resemblance to our generated texture samples, they are not the same, and visual differences are noticeable using both co-occurrence and RGB measures. 

\begin{figure}
\centering
\rotatebox{90}{\whitetxt{ssssssss}PSGAN\whitetxt{sssssss} }
\includegraphics[width=0.95\columnwidth]{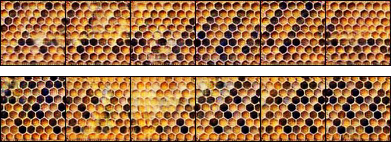} \\

\vspace{3pt}

\rotatebox{90}{\whitetxt{sss}Ours\whitetxt{sss} }
\includegraphics[width=0.95\columnwidth]{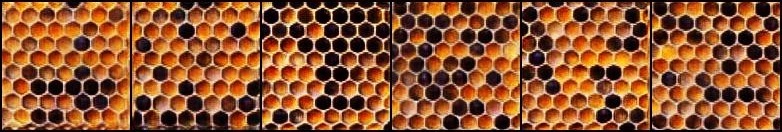} 






\caption{{\bfseries Diversity comparison.} 
Texture samples generated by PSGAN~\cite{bergmann2017learning} are demonstrated on the top rows. Several of these results seem to have a repetitive pattern and their visual diversity is limited. Our technique is conditioned on co-occurrence statistics and thus generates significantly more diverse results. 
}
\label{fig:divers_comp_supp}
\end{figure}

\begin{figure}
\begin{center}
\includegraphics[width=1.0\columnwidth]{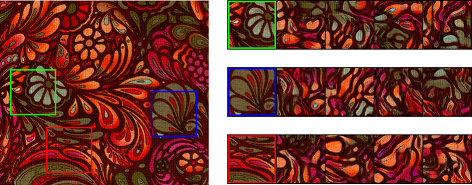}
\end{center}
\caption{{\bfseries Limitation example}. When the input texture includes elements of scale larger than our crop size of $128 \times 128$, they are not preserved in the generated samples.}
\label{fig:limitation_examples}
\end{figure}

\paragraph*{Stability of Synthesized Textures.}
We use a texture degradation test~\cite{kaspar2015self,sendik2017deep} to measure the stability of our algorithm. Specifically, we do the following: Given a co-occurrence tensor from the train set, we generate a texture sample, compute its co-occurrence tensor, and feed it back to the generator (with the same noise vector) to generate another sample. This process is repeated several times.

Figure~\ref{fig:cooc_loop} shows representative results of our stability test. As the figure illustrates, the appearance of the synthesized textures remains roughly the same. We attribute this stable behaviour to the use of the co-occurrence as a condition for the texture generation process.


To quantitatively evaluate the stability of our method, we repeated this test for the entire test set of different texture images. Following each looping iteration, we measure the $L_1$ difference between the input co-occurrence and that of the generated sample and average over all the examples in the set. For $10$ looping iterations the average $L_1$ measure remains within a $2\%$ range with respect to the average $L_1$ measure of the first iteration. This was the case on all the examined textures, indicating that the co-occurrence of the generated samples is indeed stable.



\subsection{Comparison to Previous Work} 
\label{sec:previous_work}
 We compare our method to two texture generation algorithms in the literature that are related to our work: PSGAN~\cite{bergmann2017learning} and TextureMixer~\cite{yu2019texture}. First, we examine the diversity of our synthesized texture samples against that of PSGAN, illustrated in Figure~\ref{fig:divers_comp_supp}. PSGAN is an unconditional texture synthesis algorithm. Thus, there is no control on the appearance of the generated samples, resulting in limited diversity. Our method, on the contrary, is conditioned on co-occurrence statistics that represent the texture's local properties. Different input co-occurrences result in texture samples with different appearance, which allows us to generate diverse samples.

\begin{figure*}
\centering
	\jsubfig{\includegraphics[height=1.9cm]{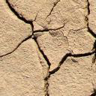}}
	{}%
	\hfill%
	\jsubfig{\includegraphics[height=1.9cm]{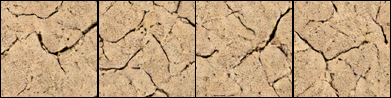}}
	{}%
	\hfill%
	\jsubfig{\includegraphics[height=1.9cm]{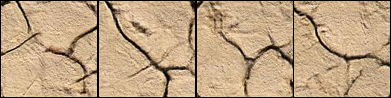}}
	{}%
	\\
	\vspace{5pt}
		\jsubfig{\includegraphics[height=1.9cm]{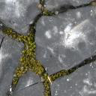}}
	{}%
	\hfill%
	\jsubfig{\includegraphics[height=1.9cm]{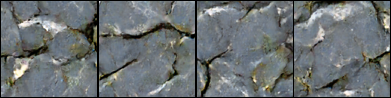}}
	{}%
	\hfill%
	\jsubfig{\includegraphics[height=1.9cm]{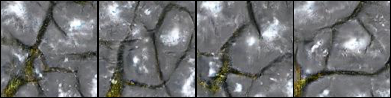}}
	{}%
	\\
	\vspace{5pt}
		\jsubfig{\includegraphics[height=1.9cm]{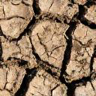}}
	{Input}%
	\hfill%
	\jsubfig{\includegraphics[height=1.9cm]{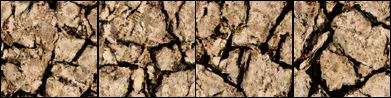}}
	{Texture Mixer}%
	\hfill%
	\jsubfig{\includegraphics[height=1.9cm]{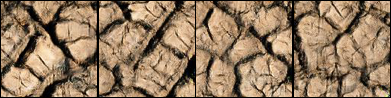}}
	{Ours}%
\caption{{\bfseries Fidelity and diversity comparison.} 
Given an input crop (left), we demonstrate samples synthesized using Texture Mixer~\cite{yu2019texture} (center) and our technique (right). Note that unlike Texture Mixer, we do not encode the input crop directly---we generate samples conditioned on its co-occurrence matrix. While both methods generate diverse results, our method seems to better respect the properties of the input texture crop.}
\label{fig:fidel_divers_comp_supp}
\end{figure*}

\begin{figure*}
\begin{center}
	\jsubfig{\includegraphics[height=4.2cm]{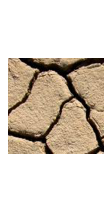}}
	{}%
	\hfill%
	\rotatebox{90}{\whitetxt{sssss}Ours\whitetxt{ssssssss}  Texture Mixer }
	\jsubfig{\includegraphics[height=4.2cm]{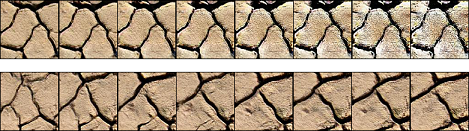}}
	{}%
\\
\vspace{8pt}
	\jsubfig{\includegraphics[height=4.22cm]{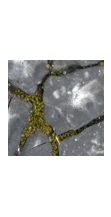}}
	{}%
	\hfill%
	\rotatebox{90}{\whitetxt{sssss}Ours\whitetxt{ssssssss}  Texture Mixer }
	\jsubfig{\includegraphics[height=4.22cm]{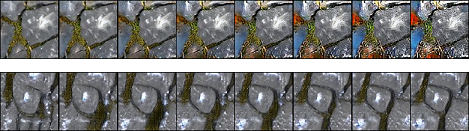}}
	{}%
	\\
	\vspace{8pt}
	\jsubfig{\includegraphics[height=4.2cm]{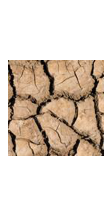}}
	{}%
	\hfill%
	\rotatebox{90}{\whitetxt{sssss}Ours\whitetxt{ssssssss}  Texture Mixer }
	\jsubfig{\includegraphics[height=4.2cm]{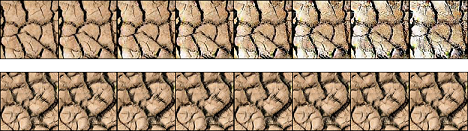}}
	{}%
\caption{{\bfseries Stability comparison.} Given an input crop (left), we iteratively generate a sample with Texture Mixer~\cite{yu2019texture} (top rows) and with our method (bottom rows). While Texture Mixer suffers from severe artifacts along the iterations, our texture synthesis method remains stable.}
\label{fig:stability_comp_supp}
\end{center}
\end{figure*}

\begin{figure*}
\begin{center}
\rotatebox{90}{\whitetxt{ssssss}Ours\whitetxt{sssssssss}  Texture Mixer \whitetxt{ssssss} $\alpha$-blend}
\includegraphics[width=0.95\textwidth]{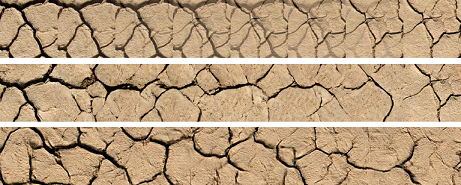} \\
\vspace{0.3cm}
\rotatebox{90}{\whitetxt{ssssss}Ours\whitetxt{sssssssss}  Texture Mixer \whitetxt{ssssss} $\alpha$-blend}
\includegraphics[width=0.95\textwidth]{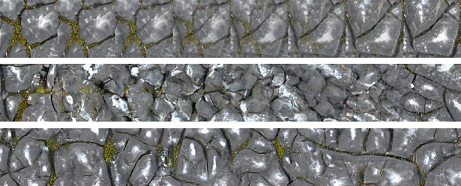} 
\caption{{\bfseries Interpolation comparison.} We compare interpolation results obtained with $\alpha$-blending (top rows), Texture Mixer~\cite{yu2019texture} (middle rows) and our method (bottom rows).
The interpolation stripes are of size $1024 \times 128$ pixels. As illustrated above, our interpolations tend to be smoother and less repetitive than those of Texture Mixer.}
\label{fig:interp_comp_supp}
\end{center}
\end{figure*}

Next, we compare our generated textures to the ones obtained with Texture Mixer~\cite{yu2019texture}. 
Our comparison focuses on three different aspects: fidelity and diversity (Figure \ref{fig:fidel_divers_comp_supp}), stability (Figure \ref{fig:stability_comp_supp}), and interpolation between texture regions with different properties (Figure \ref{fig:interp_comp_supp}). We use their publicly available model, which was trained on earth textures. 

As illustrated in Figure \ref{fig:fidel_divers_comp_supp}, while both methods generate diverse results, our method has somewhat higher fidelity to the input texture. In terms of stability, as demonstrated in Figure \ref{fig:stability_comp_supp}, their method gradually breaks, while ours remains stable. This experiment illustrates that their encoder is sensitive to the distribution of which it was trained on, and as the generated samples deviate from this distribution, it cannot accurately encode it to the latent space. Lastly, as illustrated in Figure \ref{fig:interp_comp_supp}, the local properties change more steadily and smoothly using our method.


\subsection{Ablation Study}
\label{sec:ablations}
\begin{figure*}
\begin{center}
\begin{tabular}{ c c c }

\multirow{9}{*}{
\includegraphics[width=0.3\linewidth]{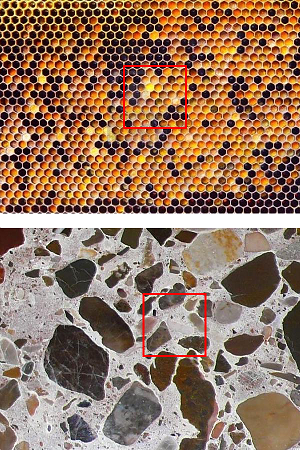}
}

&
\whitetxt{Empty line}
&
\whitetxt{Empty line}
\\

&
Without co-occurrence condition at G
&
Without co-occurrence condition at G
\\

&
\includegraphics[width=0.3\linewidth]{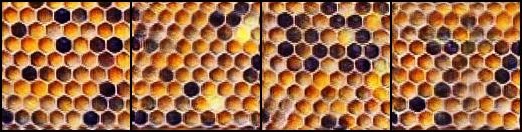}
&
\includegraphics[width=0.3\linewidth]{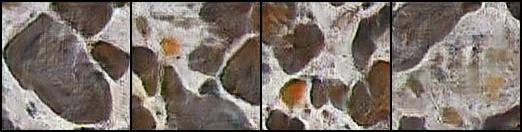}
\\

&
Without co-occurrence condition at D 
&
Without co-occurrence condition at D
\\

&
\includegraphics[width=0.3\linewidth]{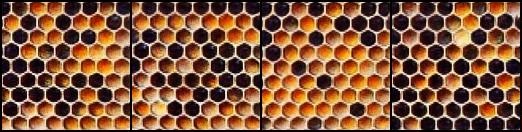}
&
\includegraphics[width=0.3\linewidth]{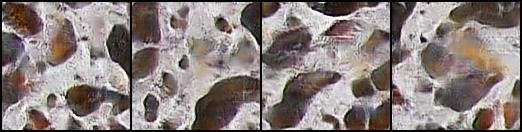}
\\

&
Without co-occurrence loss
&
Without co-occurrence loss
\\

&
\includegraphics[width=0.3\linewidth]{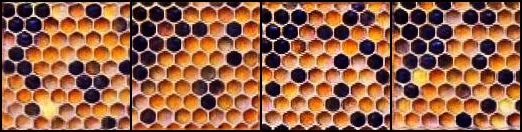}
&
\includegraphics[width=0.3\linewidth]{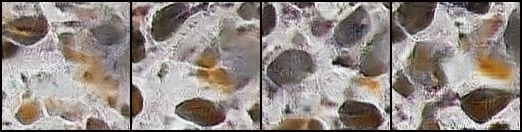}
\\

&
With all components
&
With all components
\\

&
\includegraphics[width=0.3\linewidth]{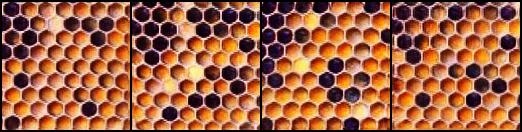}
&
\includegraphics[width=0.3\linewidth]{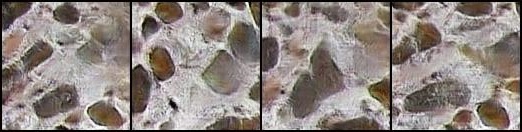}
\\

\end{tabular}
\end{center}
\caption{{\bfseries Ablation study.}  Each row on the center and right shows several synthesized texture samples, conditioned on the co-occurrence of the reference crop, marked on the texture exemplars on the left. Each time one component of our method is turned off. G and D stands for Generator and Discriminator, respectively. When one of the components is missing, the fidelity to the input co-occurrence is compromised. When all the components are used, the generated samples better respect the properties of the reference crop. 
}
\label{fig:ablation}
\end{figure*}

In order to validate the importance of the different components in our framework, we perform several experiments, omitting key components one at a time. Specifically, we train our cGAN without the co-occurrence condition at the generator or at the discriminator, or without the co-occurrence loss. Results are shown in Figure~\ref{fig:ablation}.

When the generator is non-conditional, the control over the generation process is completely lost. Omitting the co-occurrence condition form the discriminator or not using the co-occurrence loss degrades the fidelity of generated texture samples, such that their properties are different than those of the reference crop, from which the co-occurrence was collected. When utilizing all the components, our method can better preserve the local texture properties, represented by the co-occurrence statistics.


\subsection{Limitations} 
\label{sec:limitations}
The quality of our results degrade when texture elements are larger than the crop size we use (i.e., $128 \times 128$ pixels). The method also fails to capture texture characterized by global structure. We show examples of these kinds of textures in Figure \ref{fig:limitation_examples}.

\section{Applications} \label{sec:applications}

We demonstrate our co-occurrence based texture synthesis in the context of several applications: an interactive texture editing framework, dynamic texture morph generation and controllable textures. 




\begin{figure*}
\centering
	\jsubfig{\includegraphics[height=1.9cm]{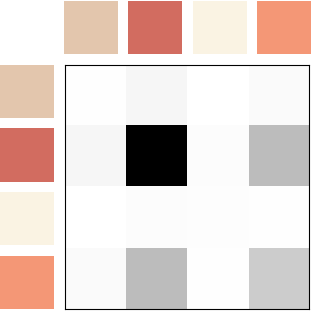}}
	{}%
	\hfill%
	\jsubfig{\includegraphics[height=1.9cm]{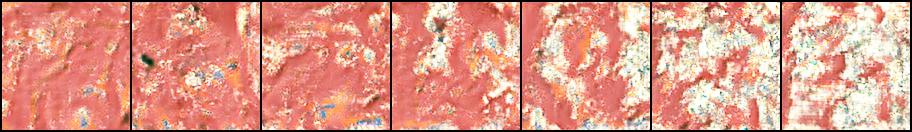}}
	{}%
	\hfill%
	\jsubfig{\includegraphics[height=1.9cm]{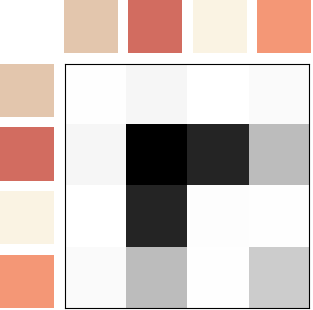}}
	{}
	\\
	 \vspace{3pt}
		\jsubfig{\includegraphics[height=1.9cm]{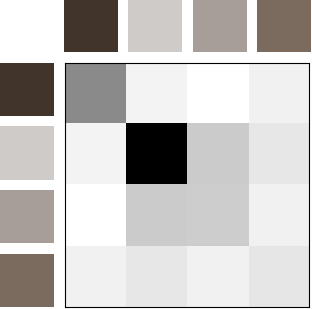}}
	{}%
	\hfill%
	\jsubfig{\includegraphics[height=1.9cm]{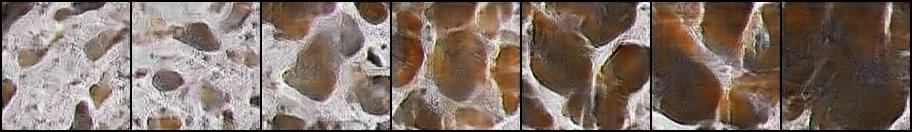}}
	{}%
	\hfill%
	\jsubfig{\includegraphics[height=1.9cm]{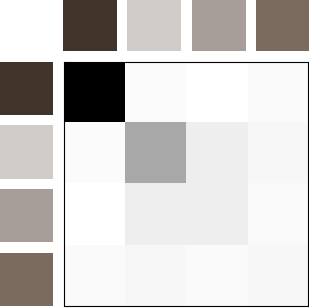}}
	{}
\caption{{\bfseries Interactive texture editing.} With our representation, a user can locally adjust the generated image by editing its corresponding co-occurrence matrix. 
On the left, we illustrate the initial result and its co-occurrence matrix. 
We demonstrate the sequence of results obtained by multiplying a selected bin with an increasing factor (and normalizing accordingly, as described in the text). The co-occurrence matrix on the right corresponds to the rightmost image.
Note that darker colors correspond to pixel values which co-occur with high probability.}
\label{fig:interactive_new}
\end{figure*}

\paragraph*{Interactive Texture Editing} \label{seq:interactive_editing}

Our technique lets users edit locally generated texture by modifying the corresponding co-occurrence matrix. 
A user can modify, for instance, a single bin $M(\tau_i, \tau_j)$ in the co-occurrence matrix. %
We can then generate a texture image with the modified co-occurrence (after normalizing the matrix to sum to $1$).
In Figure \ref{fig:interactive_new}, we demonstrate texture sequences obtained from manipulating a single bin in the co-occurrence matrix (i.e., multiplying it with an increasing factor).

\begin{figure*}
\centering
	\jsubfig{\includegraphics[height=3.0cm]{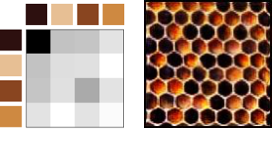}}
	{}%
	\hfill%
	\jsubfig{\includegraphics[height=3.0cm]{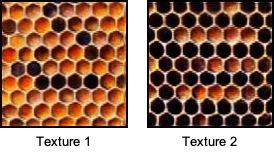}}
	{}%
	\hfill%
	\jsubfig{\includegraphics[height=3.0cm]{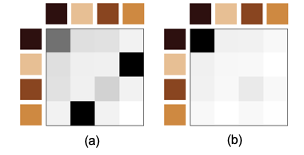}}
	{}
	\\
	\jsubfig{\includegraphics[height=3.0cm]{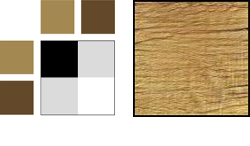}}
	{Original pair}%
	\hfill%
	\jsubfig{\includegraphics[height=3.0cm]{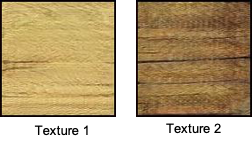}}
	{Edited images}%
	\hfill%
	\jsubfig{\includegraphics[height=3.0cm]{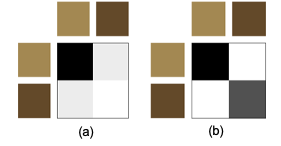}}
	{Edited co-occurrences}%
\caption{{\bfseries User study.} Given an original texture image and two different co-occurrence based edits, participants were asked to match between the edited textures (in the center) and their edited co-occurrence matrix (on the right). The correct matches are provided\protect\footnotemark. Note that darker colors correspond to pixel values which co-occur with high probability.}
\label{fig:user_study}
\end{figure*}
\footnotetext{The co-occurrence matrix corresponding to Texture 1 is (a) in both cases.}

We conducted a user study to quantitatively evaluate how interpertable and accessible our interactive texture editing is for users who are not familiar with co-occurrence statistics. Participants were first provided a brief introduction to co-occurrences using the toy example illustrated in Figure \ref{fig:cooc_matrix_example}. Then, they were presented with image triplets containing one original texture image and two edited images. 
The participants were asked to match the edited images with their co-occurrence matrices. The co-occurrence matrix corresponding to the original image was provided. Participants were also asked to rank how confident they are in their selection using a 5-point Likert scale.

Seventy two users participated in the study. Each participant was shown  image triplets generated from 5 different textures and the number of clusters $k$ was set to either $k=2$ or $k=4$. In Figure \ref{fig:user_study}, we provide a few sample questions from our user study. 

On average, $83\%$ of the time the participants successfully matched between the image and its corresponding co-occurrence matrix. The participants were confident in their selection $72\%$ of the time. For these confident selections, the success rate was $86\%$ on average. As our study illustrates, in most cases users understand how an edit in the image is reflected in the co-occurrence matrix. Therefore, directly editing the co-occurrence matrix can provide meaningful local control of the texture generation process.



\begin{figure*}
\begin{center}
\includegraphics[width=0.95\linewidth]{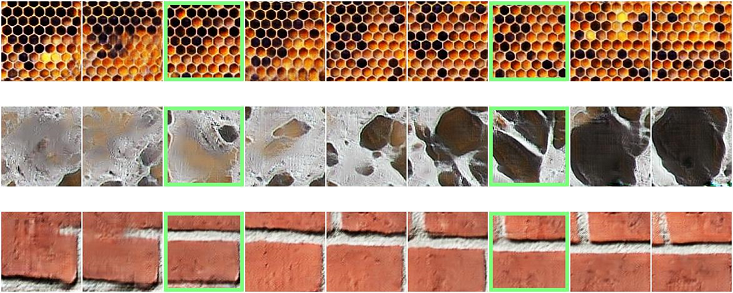} \\
\end{center}
\caption{{\bfseries Texture interpolation and extrapolation.} Examples of interpolated and extrapolated textures, generated between two sample co-occurrence vectors (corresponding to the generated images marked in green). As the figure illustrates, the extrapolated examples extend the smooth interpolation sequence, magnifying the differences in the local texture properties. Please refer to the accompanying video for the full dynamic sequences.
}
\label{fig:interp_extrap}
\end{figure*}

\paragraph*{Texture Morphing}
We generate a dynamic texture morph by interpolating and extrapolating between randomly selected co-occurrence tensors of various local texture regions. The generated image sequence, obtained by conditioning on the smooth co-occurrence sequence and a fixed random noise seed, exhibits a unique temporal behavior. In Figure \ref{fig:interp_extrap}, we illustrate a few representative frames along the sequence. We provide full dynamic sequences in the accompanying video.


\paragraph*{Large Controllable Textures} 
Our texture synthesis algorithm is fully convolutional and thus can generate arbitrarily large textures. In addition, since it is conditioned on co-occurrence, we have control over the local appearance of the synthesized texture.

In order to demonstrate this ability, we take two crops of $128 \times 128$ pixels with different properties from a texture exemplar. Then, we tile their corresponding co-occurrence tensors in a desired spatial arrangement, and run it through the generator. This way we can obtain a texture with desired appearance.



\begin{figure*}
\begin{center}
\includegraphics[width=1.0\textwidth]{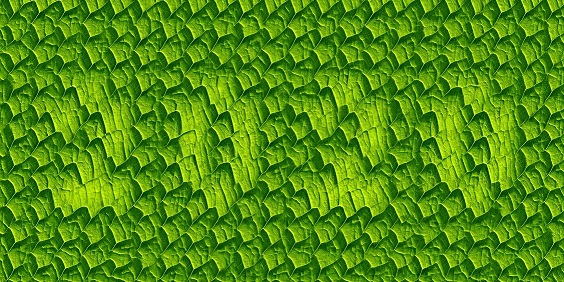}
\caption{{\bfseries Generating large texture with control.} Texture of size $2816 \times 1408$ pixels produced by our method. We can control the local appearance of the texture, by constructing the co-occurrence condition to the generator network. For this example, we used co-occurrences from \textit{only} two $128 \times 128$ crops of the original texture!}
\label{fig:large_supp}
\end{center}
\end{figure*}

Figure~\ref{fig:large_supp} shows a large synthesized texture in that manner. The fidelity to the co-occurrence condition enables us to depict the number "2020". The diversity property of our algorithm enables producing this large size texture with a plausible appearance.

\section{Conclusion}
We proposed a co-occurrence based texture synthesis method, where local texture properties are captured by a co-occurrence tensor. While the computation of the co-occurrence for a given texture crop is deterministic, the opposite direction is not. That is, different texture crops can have similar co-occurrence statistics. As we show throughout our work, our fully convolutional cGAN managed to learn an accurate and stable non-deterministic mapping from the co-occurrence space back to the image space. 


We believe that co-occurrences strikes the right balance between non-parametric methods that are difficult to control and manipulate and parametric methods that may have limited expressive power. 
Generally speaking, we hope to explore and learn about other deep frameworks where powerful traditional tools are integrated into neural networks to advance their respective fields.

%
%
\bibliographystyle{ACM-Reference-Format}
\bibliography{references}


\begin{thebibliography}{35}


\ifx \showCODEN    \undefined \def \showCODEN     #1{\unskip}     \fi
\ifx \showDOI      \undefined \def \showDOI       #1{#1}\fi
\ifx \showISBNx    \undefined \def \showISBNx     #1{\unskip}     \fi
\ifx \showISBNxiii \undefined \def \showISBNxiii  #1{\unskip}     \fi
\ifx \showISSN     \undefined \def \showISSN      #1{\unskip}     \fi
\ifx \showLCCN     \undefined \def \showLCCN      #1{\unskip}     \fi
\ifx \shownote     \undefined \def \shownote      #1{#1}          \fi
\ifx \showarticletitle \undefined \def \showarticletitle #1{#1}   \fi
\ifx \showURL      \undefined \def \showURL       {\relax}        \fi
\providecommand\bibfield[2]{#2}
\providecommand\bibinfo[2]{#2}
\providecommand\natexlab[1]{#1}
\providecommand\showeprint[2][]{arXiv:#2}

\bibitem[\protect\citeauthoryear{Bau, Zhu, Strobelt, Zhou, Tenenbaum, Freeman,
  and Torralba}{Bau et~al\mbox{.}}{2019}]%
        {bau2019visualizing}
\bibfield{author}{\bibinfo{person}{David Bau}, \bibinfo{person}{Jun-Yan Zhu},
  \bibinfo{person}{Hendrik Strobelt}, \bibinfo{person}{Bolei Zhou},
  \bibinfo{person}{Joshua~B Tenenbaum}, \bibinfo{person}{William~T Freeman},
  {and} \bibinfo{person}{Antonio Torralba}.} \bibinfo{year}{2019}\natexlab{}.
\newblock \showarticletitle{Visualizing and understanding generative
  adversarial networks}.
\newblock \bibinfo{journal}{\emph{arXiv preprint arXiv:1901.09887}}
  (\bibinfo{year}{2019}).
\newblock


\bibitem[\protect\citeauthoryear{Bellini, Kleiman, and Cohen-Or}{Bellini
  et~al\mbox{.}}{2016}]%
        {bellini2016time}
\bibfield{author}{\bibinfo{person}{Rachele Bellini}, \bibinfo{person}{Yanir
  Kleiman}, {and} \bibinfo{person}{Daniel Cohen-Or}.}
  \bibinfo{year}{2016}\natexlab{}.
\newblock \showarticletitle{Time-varying weathering in texture space}.
\newblock \bibinfo{journal}{\emph{ACM Transactions on Graphics (TOG)}}
  \bibinfo{volume}{35}, \bibinfo{number}{4} (\bibinfo{year}{2016}),
  \bibinfo{pages}{141}.
\newblock


\bibitem[\protect\citeauthoryear{Bergmann, Jetchev, and Vollgraf}{Bergmann
  et~al\mbox{.}}{2017}]%
        {bergmann2017learning}
\bibfield{author}{\bibinfo{person}{Urs Bergmann}, \bibinfo{person}{Nikolay
  Jetchev}, {and} \bibinfo{person}{Roland Vollgraf}.}
  \bibinfo{year}{2017}\natexlab{}.
\newblock \showarticletitle{Learning texture manifolds with the periodic
  spatial GAN}.
\newblock \bibinfo{journal}{\emph{arXiv preprint arXiv:1705.06566}}
  (\bibinfo{year}{2017}).
\newblock


\bibitem[\protect\citeauthoryear{Bonet}{Bonet}{1997}]%
        {Bonet97}
\bibfield{author}{\bibinfo{person}{Jeremy S.~De Bonet}.}
  \bibinfo{year}{1997}\natexlab{}.
\newblock \showarticletitle{Multiresolution sampling procedure for analysis and
  synthesis of texture images}. In \bibinfo{booktitle}{\emph{Proceedings of the
  24th Annual Conference on Computer Graphics and Interactive Techniques,
  {SIGGRAPH} 1997, Los Angeles, CA, USA, August 3-8, 1997}}.
  \bibinfo{pages}{361--368}.
\newblock
\urldef\tempurl%
\url{https://doi.org/10.1145/258734.258882}
\showDOI{\tempurl}


\bibitem[\protect\citeauthoryear{Cimpoi, Maji, Kokkinos, Mohamed, and
  Vedaldi}{Cimpoi et~al\mbox{.}}{2014}]%
        {cimpoi2014describing}
\bibfield{author}{\bibinfo{person}{Mircea Cimpoi}, \bibinfo{person}{Subhransu
  Maji}, \bibinfo{person}{Iasonas Kokkinos}, \bibinfo{person}{Sammy Mohamed},
  {and} \bibinfo{person}{Andrea Vedaldi}.} \bibinfo{year}{2014}\natexlab{}.
\newblock \showarticletitle{Describing Textures in the Wild}. In
  \bibinfo{booktitle}{\emph{Proceedings of the IEEE Conference on Computer
  Vision and Pattern Recognition (CVPR)}}.
\newblock


\bibitem[\protect\citeauthoryear{Darabi, Shechtman, Barnes, Goldman, and
  Sen}{Darabi et~al\mbox{.}}{2012}]%
        {Darabi12:ImageMelding12}
\bibfield{author}{\bibinfo{person}{Soheil Darabi}, \bibinfo{person}{Eli
  Shechtman}, \bibinfo{person}{Connelly Barnes}, \bibinfo{person}{Dan~B.
  Goldman}, {and} \bibinfo{person}{Pradeep Sen}.}
  \bibinfo{year}{2012}\natexlab{}.
\newblock \showarticletitle{{I}mage {M}elding: Combining Inconsistent Images
  using Patch-based Synthesis}.
\newblock \bibinfo{journal}{\emph{ACM Transactions on Graphics (TOG)
  (Proceedings of SIGGRAPH 2012)}} \bibinfo{volume}{31}, \bibinfo{number}{4},
  Article \bibinfo{articleno}{82} (\bibinfo{year}{2012}),
  \bibinfo{numpages}{82:1--82:10}~pages.
\newblock


\bibitem[\protect\citeauthoryear{Efros and Freeman}{Efros and Freeman}{2001}]%
        {efros2001image}
\bibfield{author}{\bibinfo{person}{Alexei~A Efros} {and}
  \bibinfo{person}{William~T Freeman}.} \bibinfo{year}{2001}\natexlab{}.
\newblock \showarticletitle{Image quilting for texture synthesis and transfer}.
  In \bibinfo{booktitle}{\emph{Proceedings of the 28th annual conference on
  Computer graphics and interactive techniques}}. ACM,
  \bibinfo{pages}{341--346}.
\newblock


\bibitem[\protect\citeauthoryear{Fr{\"u}hst{\"u}ck, Alhashim, and
  Wonka}{Fr{\"u}hst{\"u}ck et~al\mbox{.}}{2019}]%
        {fruhstuck2019tilegan}
\bibfield{author}{\bibinfo{person}{Anna Fr{\"u}hst{\"u}ck},
  \bibinfo{person}{Ibraheem Alhashim}, {and} \bibinfo{person}{Peter Wonka}.}
  \bibinfo{year}{2019}\natexlab{}.
\newblock \showarticletitle{TileGAN: synthesis of large-scale non-homogeneous
  textures}.
\newblock \bibinfo{journal}{\emph{ACM Transactions on Graphics (TOG)}}
  \bibinfo{volume}{38}, \bibinfo{number}{4} (\bibinfo{year}{2019}),
  \bibinfo{pages}{1--11}.
\newblock


\bibitem[\protect\citeauthoryear{Gatys, Ecker, and Bethge}{Gatys
  et~al\mbox{.}}{2015}]%
        {gatys2015texture}
\bibfield{author}{\bibinfo{person}{Leon Gatys}, \bibinfo{person}{Alexander~S.
  Ecker}, {and} \bibinfo{person}{Matthias Bethge}.}
  \bibinfo{year}{2015}\natexlab{}.
\newblock \showarticletitle{Texture synthesis using convolutional neural
  networks}. In \bibinfo{booktitle}{\emph{Advances in neural information
  processing systems}}. \bibinfo{pages}{262--270}.
\newblock


\bibitem[\protect\citeauthoryear{Gulrajani, Ahmed, Arjovsky, Dumoulin, and
  Courville}{Gulrajani et~al\mbox{.}}{2017}]%
        {gulrajani2017improved}
\bibfield{author}{\bibinfo{person}{Ishaan Gulrajani}, \bibinfo{person}{Faruk
  Ahmed}, \bibinfo{person}{Mart{\'i}n Arjovsky}, \bibinfo{person}{Vincent
  Dumoulin}, {and} \bibinfo{person}{Aaron~C. Courville}.}
  \bibinfo{year}{2017}\natexlab{}.
\newblock \showarticletitle{Improved Training of Wasserstein GANs}. In
  \bibinfo{booktitle}{\emph{Proceesings of the Annual Conference on Neural
  Information Processing Systems (NIPS)}}.
\newblock


\bibitem[\protect\citeauthoryear{Haralick, Shanmugam, and Dinstein}{Haralick
  et~al\mbox{.}}{1973}]%
        {HaralickSD73}
\bibfield{author}{\bibinfo{person}{Robert~M. Haralick}, \bibinfo{person}{K.~Sam
  Shanmugam}, {and} \bibinfo{person}{Its'hak Dinstein}.}
  \bibinfo{year}{1973}\natexlab{}.
\newblock \showarticletitle{Textural Features for Image Classification}.
\newblock \bibinfo{journal}{\emph{{IEEE} Trans. Systems, Man, and Cybernetics}}
  \bibinfo{volume}{3}, \bibinfo{number}{6} (\bibinfo{year}{1973}),
  \bibinfo{pages}{610--621}.
\newblock


\bibitem[\protect\citeauthoryear{Heeger and Bergen}{Heeger and Bergen}{1995}]%
        {Heeger1995}
\bibfield{author}{\bibinfo{person}{David~J. Heeger} {and}
  \bibinfo{person}{James~R. Bergen}.} \bibinfo{year}{1995}\natexlab{}.
\newblock \showarticletitle{Pyramid-based Texture Analysis/Synthesis}. In
  \bibinfo{booktitle}{\emph{Proceedings of the 22Nd Annual Conference on
  Computer Graphics and Interactive Techniques}}
  \emph{(\bibinfo{series}{SIGGRAPH '95})}. \bibinfo{publisher}{ACM},
  \bibinfo{address}{New York, NY, USA}, \bibinfo{pages}{229--238}.
\newblock
\showISBNx{0-89791-701-4}
\urldef\tempurl%
\url{https://doi.org/10.1145/218380.218446}
\showDOI{\tempurl}


\bibitem[\protect\citeauthoryear{Isola, Zoran, Krishnan, and Adelson}{Isola
  et~al\mbox{.}}{2014}]%
        {IsolaZKA14}
\bibfield{author}{\bibinfo{person}{Phillip Isola}, \bibinfo{person}{Daniel
  Zoran}, \bibinfo{person}{Dilip Krishnan}, {and} \bibinfo{person}{Edward~H.
  Adelson}.} \bibinfo{year}{2014}\natexlab{}.
\newblock \showarticletitle{Crisp Boundary Detection Using Pointwise Mutual
  Information}. In \bibinfo{booktitle}{\emph{Computer Vision - {ECCV} 2014 -
  13th European Conference, Zurich, Switzerland, September 6-12, 2014,
  Proceedings, Part {III}}}. \bibinfo{pages}{799--814}.
\newblock


\bibitem[\protect\citeauthoryear{Jetchev, Bergmann, and Vollgraf}{Jetchev
  et~al\mbox{.}}{2016}]%
        {jetchev2016texture}
\bibfield{author}{\bibinfo{person}{Nikolay Jetchev}, \bibinfo{person}{Urs
  Bergmann}, {and} \bibinfo{person}{Roland Vollgraf}.}
  \bibinfo{year}{2016}\natexlab{}.
\newblock \showarticletitle{Texture synthesis with spatial generative
  adversarial networks}.
\newblock \bibinfo{journal}{\emph{Workshop on Adversarial Training, NIPS}}
  (\bibinfo{year}{2016}).
\newblock


\bibitem[\protect\citeauthoryear{Jevnisek and Avidan}{Jevnisek and
  Avidan}{2017}]%
        {jevnisek2017cooc}
\bibfield{author}{\bibinfo{person}{Roy~J. Jevnisek} {and} \bibinfo{person}{Shai
  Avidan}.} \bibinfo{year}{2017}\natexlab{}.
\newblock \showarticletitle{Co-occurrence Filter}. In
  \bibinfo{booktitle}{\emph{Proceedings of the IEEE Conference on Computer
  Vision and Pattern Recognition (CVPR)}}. \bibinfo{pages}{3184--3192}.
\newblock


\bibitem[\protect\citeauthoryear{Julesz}{Julesz}{1962}]%
        {Julesz:62}
\bibfield{author}{\bibinfo{person}{Bela Julesz}.}
  \bibinfo{year}{1962}\natexlab{}.
\newblock \showarticletitle{Visual Pattern Discrimination}.
\newblock \bibinfo{journal}{\emph{Information Theory, IRE Transactions on}}
  \bibinfo{volume}{8} (\bibinfo{date}{03} \bibinfo{year}{1962}),
  \bibinfo{pages}{84 -- 92}.
\newblock
\urldef\tempurl%
\url{https://doi.org/10.1109/TIT.1962.1057698}
\showDOI{\tempurl}


\bibitem[\protect\citeauthoryear{Kaspar, Neubert, Lischinski, Pauly, and
  Kopf}{Kaspar et~al\mbox{.}}{2015}]%
        {kaspar2015self}
\bibfield{author}{\bibinfo{person}{lexandre Kaspar}, \bibinfo{person}{Boris
  Neubert}, \bibinfo{person}{Dani Lischinski}, \bibinfo{person}{Mark Pauly},
  {and} \bibinfo{person}{Johannes Kopf}.} \bibinfo{year}{2015}\natexlab{}.
\newblock \showarticletitle{Self tuning texture optimization}.
\newblock \bibinfo{journal}{\emph{Computer Graphics Forum}}
  \bibinfo{volume}{34}, \bibinfo{number}{2} (\bibinfo{year}{2015}),
  \bibinfo{pages}{349--359}.
\newblock


\bibitem[\protect\citeauthoryear{Kat, Jevnisek, and Avidan}{Kat
  et~al\mbox{.}}{2018}]%
        {kat2018matching}
\bibfield{author}{\bibinfo{person}{Rotal Kat}, \bibinfo{person}{Roy Jevnisek},
  {and} \bibinfo{person}{Shai Avidan}.} \bibinfo{year}{2018}\natexlab{}.
\newblock \showarticletitle{Matching Pixels using Co-Occurrence Statistics}. In
  \bibinfo{booktitle}{\emph{Proceedings of the IEEE Conference on Computer
  Vision and Pattern Recognition (CVPR)}}. \bibinfo{pages}{1751--1759}.
\newblock


\bibitem[\protect\citeauthoryear{Kwatra, Essa, Bobick, and Kwatra}{Kwatra
  et~al\mbox{.}}{2005}]%
        {kwatra2005texture}
\bibfield{author}{\bibinfo{person}{Vivek Kwatra}, \bibinfo{person}{Irfan Essa},
  \bibinfo{person}{Aaron Bobick}, {and} \bibinfo{person}{Nipun Kwatra}.}
  \bibinfo{year}{2005}\natexlab{}.
\newblock \showarticletitle{Texture optimization for example-based synthesis}.
\newblock \bibinfo{journal}{\emph{ACM Transactions on Graphics (ToG)}}
  \bibinfo{volume}{24}, \bibinfo{number}{3} (\bibinfo{year}{2005}),
  \bibinfo{pages}{795--802}.
\newblock


\bibitem[\protect\citeauthoryear{Kwatra, Sch{\"o}dl, Essa, Turk, and
  Bobick}{Kwatra et~al\mbox{.}}{2003}]%
        {kwatra2003graphcut}
\bibfield{author}{\bibinfo{person}{Vivek Kwatra}, \bibinfo{person}{Arno
  Sch{\"o}dl}, \bibinfo{person}{Irfan Essa}, \bibinfo{person}{Greg Turk}, {and}
  \bibinfo{person}{Aaron Bobick}.} \bibinfo{year}{2003}\natexlab{}.
\newblock \showarticletitle{Graphcut textures: image and video synthesis using
  graph cuts}.
\newblock \bibinfo{journal}{\emph{ACM Transactions on Graphics (ToG)}}
  \bibinfo{volume}{22}, \bibinfo{number}{3} (\bibinfo{year}{2003}),
  \bibinfo{pages}{277--286}.
\newblock


\bibitem[\protect\citeauthoryear{Li and Wand}{Li and Wand}{2016}]%
        {li2016precomputed}
\bibfield{author}{\bibinfo{person}{Chuan Li} {and} \bibinfo{person}{Michael
  Wand}.} \bibinfo{year}{2016}\natexlab{}.
\newblock \showarticletitle{Precomputed real-time texture synthesis with
  markovian generative adversarial networks}. In
  \bibinfo{booktitle}{\emph{European Conference on Computer Vision}}. Springer,
  \bibinfo{pages}{702--716}.
\newblock


\bibitem[\protect\citeauthoryear{Li, Fang, Yang, Wang, Lu, and Yang}{Li
  et~al\mbox{.}}{2017}]%
        {Li2017:DiversifiedTexture}
\bibfield{author}{\bibinfo{person}{Yijun Li}, \bibinfo{person}{Chen Fang},
  \bibinfo{person}{Jimei Yang}, \bibinfo{person}{Zhaowen Wang},
  \bibinfo{person}{Xin Lu}, {and} \bibinfo{person}{Ming-Hsuan Yang}.}
  \bibinfo{year}{2017}\natexlab{}.
\newblock \showarticletitle{Diversified Texture Synthesis with Feed-Forward
  Networks}. In \bibinfo{booktitle}{\emph{Proceedings of the IEEE Conference on
  Computer Vision and Pattern Recognition (CVPR)}}. \bibinfo{pages}{266--274}.
\newblock
\urldef\tempurl%
\url{https://doi.org/10.1109/CVPR.2017.36}
\showDOI{\tempurl}


\bibitem[\protect\citeauthoryear{Liu, Gousseau, and Xia}{Liu
  et~al\mbox{.}}{2016}]%
        {liu2016texture}
\bibfield{author}{\bibinfo{person}{Gang Liu}, \bibinfo{person}{Yann Gousseau},
  {and} \bibinfo{person}{Gui-Song Xia}.} \bibinfo{year}{2016}\natexlab{}.
\newblock \showarticletitle{Texture synthesis through convolutional neural
  networks and spectrum constraints}. In \bibinfo{booktitle}{\emph{23rd
  International Conference on Pattern Recognition (ICPR)}}. IEEE,
  \bibinfo{pages}{3234--3239}.
\newblock


\bibitem[\protect\citeauthoryear{Matusik, Zwicker, and Durand}{Matusik
  et~al\mbox{.}}{2005}]%
        {Matusik:2005:TDU}
\bibfield{author}{\bibinfo{person}{Wojciech Matusik}, \bibinfo{person}{Matthias
  Zwicker}, {and} \bibinfo{person}{Fr{\'e}do Durand}.}
  \bibinfo{year}{2005}\natexlab{}.
\newblock \showarticletitle{Texture Design Using a Simplicial Complex of
  Morphable Textures}.
\newblock \bibinfo{journal}{\emph{ACM Trans. Graph.}} \bibinfo{volume}{24},
  \bibinfo{number}{3} (\bibinfo{date}{July} \bibinfo{year}{2005}),
  \bibinfo{pages}{787--794}.
\newblock
\showISSN{0730-0301}
\urldef\tempurl%
\url{https://doi.org/10.1145/1073204.1073262}
\showDOI{\tempurl}


\bibitem[\protect\citeauthoryear{Portilla and Simoncelli}{Portilla and
  Simoncelli}{2000}]%
        {portilla2000parametric}
\bibfield{author}{\bibinfo{person}{Javier Portilla} {and}
  \bibinfo{person}{Eero~P Simoncelli}.} \bibinfo{year}{2000}\natexlab{}.
\newblock \showarticletitle{A parametric texture model based on joint
  statistics of complex wavelet coefficients}.
\newblock \bibinfo{journal}{\emph{International journal of computer vision}}
  \bibinfo{volume}{40}, \bibinfo{number}{1} (\bibinfo{year}{2000}),
  \bibinfo{pages}{49--70}.
\newblock


\bibitem[\protect\citeauthoryear{Rabin, Peyr{\'e}, Delon, and Bernot}{Rabin
  et~al\mbox{.}}{2012}]%
        {Rabin2010}
\bibfield{author}{\bibinfo{person}{Julien Rabin}, \bibinfo{person}{Gabriel
  Peyr{\'e}}, \bibinfo{person}{Julie Delon}, {and} \bibinfo{person}{Marc
  Bernot}.} \bibinfo{year}{2012}\natexlab{}.
\newblock \showarticletitle{Wasserstein Barycenter and Its Application to
  Texture Mixing}. In \bibinfo{booktitle}{\emph{Scale Space and Variational
  Methods in Computer Vision}}, \bibfield{editor}{\bibinfo{person}{Alfred~M.
  Bruckstein}, \bibinfo{person}{Bart~M. ter Haar~Romeny},
  \bibinfo{person}{Alexander~M. Bronstein}, {and} \bibinfo{person}{Michael~M.
  Bronstein}} (Eds.). \bibinfo{publisher}{Springer Berlin Heidelberg},
  \bibinfo{address}{Berlin, Heidelberg}, \bibinfo{pages}{435--446}.
\newblock
\showISBNx{978-3-642-24785-9}


\bibitem[\protect\citeauthoryear{Radford, Metz, and Chintala}{Radford
  et~al\mbox{.}}{2015}]%
        {radford2015unsupervised}
\bibfield{author}{\bibinfo{person}{Alec Radford}, \bibinfo{person}{Luke Metz},
  {and} \bibinfo{person}{Soumith Chintala}.} \bibinfo{year}{2015}\natexlab{}.
\newblock \showarticletitle{Unsupervised representation learning with deep
  convolutional generative adversarial networks}.
\newblock \bibinfo{journal}{\emph{arXiv preprint arXiv:1511.06434}}
  (\bibinfo{year}{2015}).
\newblock


\bibitem[\protect\citeauthoryear{Sendik and Cohen-Or}{Sendik and
  Cohen-Or}{2017}]%
        {sendik2017deep}
\bibfield{author}{\bibinfo{person}{Omry Sendik} {and} \bibinfo{person}{Daniel
  Cohen-Or}.} \bibinfo{year}{2017}\natexlab{}.
\newblock \showarticletitle{Deep correlations for texture synthesis}.
\newblock \bibinfo{journal}{\emph{ACM Transactions on Graphics (TOG)}}
  \bibinfo{volume}{36}, \bibinfo{number}{5} (\bibinfo{year}{2017}),
  \bibinfo{pages}{161}.
\newblock


\bibitem[\protect\citeauthoryear{Shen, Gu, Tang, and Zhou}{Shen
  et~al\mbox{.}}{2019}]%
        {shen2019interpreting}
\bibfield{author}{\bibinfo{person}{Yujun Shen}, \bibinfo{person}{Jinjin Gu},
  \bibinfo{person}{Xiaoou Tang}, {and} \bibinfo{person}{Bolei Zhou}.}
  \bibinfo{year}{2019}\natexlab{}.
\newblock \showarticletitle{Interpreting the latent space of gans for semantic
  face editing}.
\newblock \bibinfo{journal}{\emph{arXiv preprint arXiv:1907.10786}}
  (\bibinfo{year}{2019}).
\newblock


\bibitem[\protect\citeauthoryear{Simakov, Caspi, Shechtman, and Irani}{Simakov
  et~al\mbox{.}}{2008}]%
        {SimakovCSI08}
\bibfield{author}{\bibinfo{person}{Denis Simakov}, \bibinfo{person}{Yaron
  Caspi}, \bibinfo{person}{Eli Shechtman}, {and} \bibinfo{person}{Michal
  Irani}.} \bibinfo{year}{2008}\natexlab{}.
\newblock \showarticletitle{Summarizing visual data using bidirectional
  similarity}. In \bibinfo{booktitle}{\emph{2008 {IEEE} Computer Society
  Conference on Computer Vision and Pattern Recognition ({CVPR} 2008), 24-26
  June 2008, Anchorage, Alaska, {USA}}}.
\newblock


\bibitem[\protect\citeauthoryear{Ulyanov, Lebedev, Vedaldi, and
  Lempitsky}{Ulyanov et~al\mbox{.}}{2016}]%
        {ulyanov2016texture}
\bibfield{author}{\bibinfo{person}{Dmitry Ulyanov}, \bibinfo{person}{Vadim
  Lebedev}, \bibinfo{person}{Andrea Vedaldi}, {and} \bibinfo{person}{Victor~S.
  Lempitsky}.} \bibinfo{year}{2016}\natexlab{}.
\newblock \showarticletitle{Texture Networks: Feed-forward Synthesis of
  Textures and Stylized Images}. In \bibinfo{booktitle}{\emph{Proceedings of
  the International Conference on Machine Learning (ICML)}}.
  \bibinfo{pages}{1349--1357}.
\newblock


\bibitem[\protect\citeauthoryear{Wexler, Shechtman, and Irani}{Wexler
  et~al\mbox{.}}{2007}]%
        {wexler2007space}
\bibfield{author}{\bibinfo{person}{Yonatan Wexler}, \bibinfo{person}{Eli
  Shechtman}, {and} \bibinfo{person}{Michal Irani}.}
  \bibinfo{year}{2007}\natexlab{}.
\newblock \showarticletitle{Space-time completion of video}.
\newblock \bibinfo{journal}{\emph{IEEE Transactions on Pattern Analysis and
  Machine Intelligence}} \bibinfo{volume}{29}, \bibinfo{number}{3}
  (\bibinfo{year}{2007}), \bibinfo{pages}{463--476}.
\newblock


\bibitem[\protect\citeauthoryear{Yellott}{Yellott}{1993}]%
        {Yellott:93}
\bibfield{author}{\bibinfo{person}{John~I. Yellott}.}
  \bibinfo{year}{1993}\natexlab{}.
\newblock \showarticletitle{Implications of triple correlation uniqueness for
  texture statistics and the Julesz conjecture}.
\newblock \bibinfo{journal}{\emph{J. Opt. Soc. Am. A}} \bibinfo{volume}{10},
  \bibinfo{number}{5} (\bibinfo{date}{May} \bibinfo{year}{1993}),
  \bibinfo{pages}{777--793}.
\newblock
\urldef\tempurl%
\url{https://doi.org/10.1364/JOSAA.10.000777}
\showDOI{\tempurl}


\bibitem[\protect\citeauthoryear{Yu, Barnes, Shechtman, Amirghodsi, and
  Lukac}{Yu et~al\mbox{.}}{2019}]%
        {yu2019texture}
\bibfield{author}{\bibinfo{person}{Ning Yu}, \bibinfo{person}{Connelly Barnes},
  \bibinfo{person}{Eli Shechtman}, \bibinfo{person}{Sohrab Amirghodsi}, {and}
  \bibinfo{person}{Michal Lukac}.} \bibinfo{year}{2019}\natexlab{}.
\newblock \showarticletitle{Texture Mixer: A Network for Controllable Synthesis
  and Interpolation of Texture}.
\newblock \bibinfo{journal}{\emph{arXiv preprint arXiv:1901.03447}}
  (\bibinfo{year}{2019}).
\newblock


\bibitem[\protect\citeauthoryear{Zhou, Zhu, Bai, Lischinski, Cohen-Or, and
  Huang}{Zhou et~al\mbox{.}}{2018}]%
        {zhou2018non}
\bibfield{author}{\bibinfo{person}{Yang Zhou}, \bibinfo{person}{Zhen Zhu},
  \bibinfo{person}{Xiang Bai}, \bibinfo{person}{Dani Lischinski},
  \bibinfo{person}{Daniel Cohen-Or}, {and} \bibinfo{person}{Hui Huang}.}
  \bibinfo{year}{2018}\natexlab{}.
\newblock \showarticletitle{Non-Stationary Texture Synthesis by Adversarial
  Expansion}.
\newblock \bibinfo{journal}{\emph{ACM Transactions on Graphics (TOG)}}
  \bibinfo{volume}{37}, \bibinfo{number}{4} (\bibinfo{year}{2018}).
\newblock


\end{thebibliography}

\clearpage

\appendix
\section*{Supplementary}
In the following sections we provide additional results of our texture synthesis method.

\section{Fidelity and Diversity} \label{sec:FidelityDiversitySupp}
\begin{figure*}
\begin{center}
\includegraphics[height=5cm]{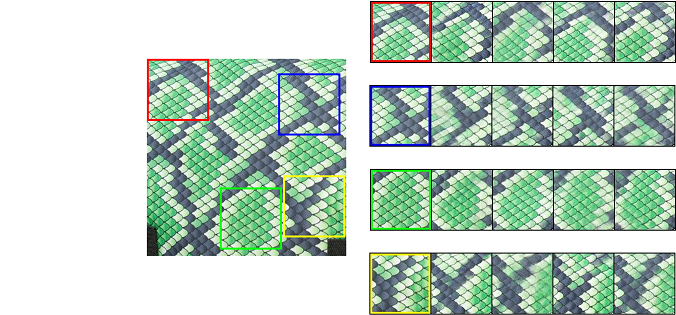} \\
\vspace{0.5cm}
\includegraphics[height=5cm]{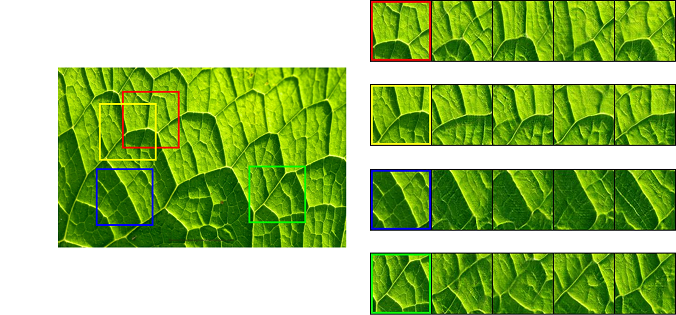} \\
\vspace{0.5cm}
\includegraphics[height=5cm]{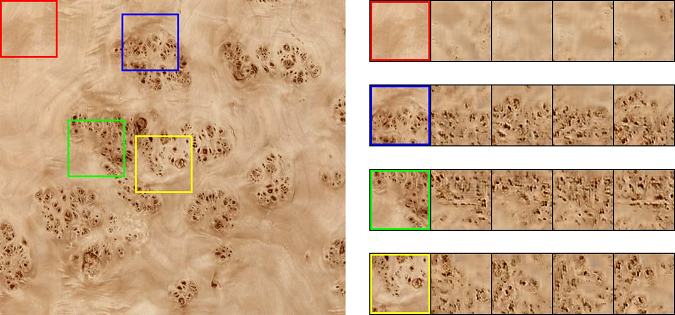} \\
\caption{{\bfseries Examples of generated texture crops from different co-occurrence tensors and noise vectors.} Note that all these conditional co-occurrence were taken from the test set, unseen during training. The synthesized texture samples from the same co-occurrence condition and different noise seeds differ from each other, while respecting the properties of the corresponding crop from the texture image.}
\label{fig:additional_fidelity_diversty}
\end{center}
\end{figure*}

Figure~\ref{fig:additional_fidelity_diversty} presents additional results, demonstrating the fidelity and diversity of the generated texture samples. See Section 4.2 in the manuscript for more details.

\section{Co-occurrence Editing} \label{sec:CoocEditSupp}

\begin{figure*}
\begin{center}
\includegraphics[width=1.0\textwidth]{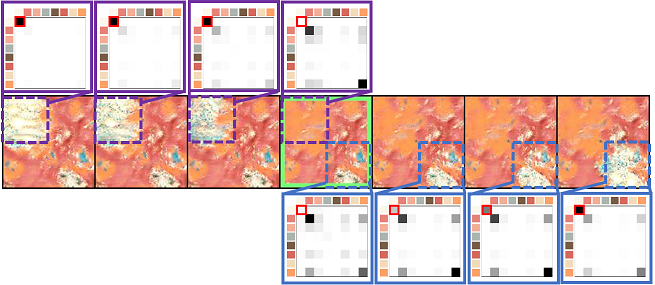} \\
\vspace{0.5cm}
\includegraphics[width=1.0\textwidth]{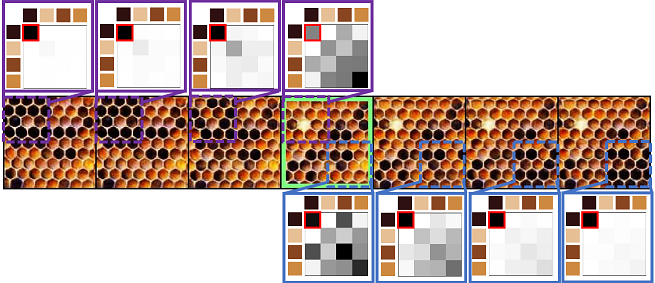} \\
\caption{
{\bfseries Additional interactive co-occurrence editing results.} Above we demonstrate how our generated textures can be edited locally. 
In the center, we illustrate the initial result and two selected regions and their co-occurrence matrices. 
We demonstrate the results obtained by multiplying a selected bin with an increasing factor to the left (editing the co-occurrence matrix inside the purple box) and to the right (editing the co-occurrence matrix inside the blue box). Darker colors in the co-occurrence matrix correspond to pixel values with co-occur with high probability.
As the figure illustrates, the edits affect the image locally. 
}
\label{fig:interactive_supp}
\end{center}
\end{figure*}

In Section 5 in the main body, we showed how to manipulate texture appearance by editing the input co-occurrence condition to the generator. Here we present additional examples of this application in Figure~\ref{fig:interactive_supp}.

\section{Comparing to tiling texture crops}
\begin{figure*}
\begin{center}
     \includegraphics[width=1.0\linewidth]{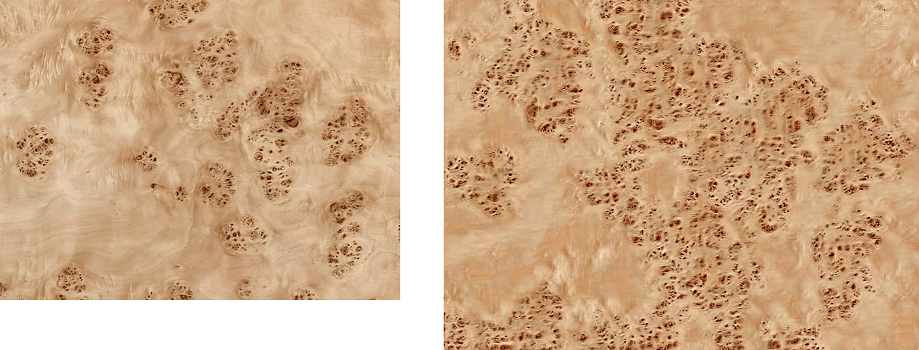} \\

     \vspace{10pt}
    
    \includegraphics[width=0.24\linewidth]{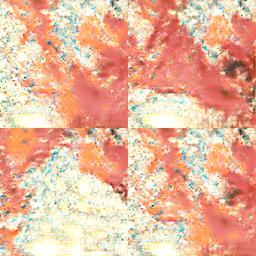}
    \hspace{1pt}
    \includegraphics[width=0.24\linewidth]{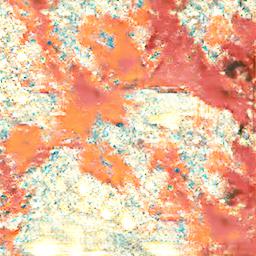}
    \hspace{6pt}
    \includegraphics[width=0.24\linewidth]{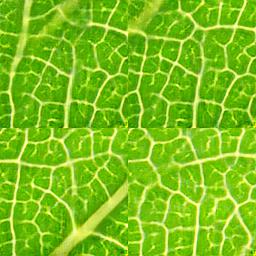}
    \hspace{1pt}
    \includegraphics[width=0.24\linewidth]{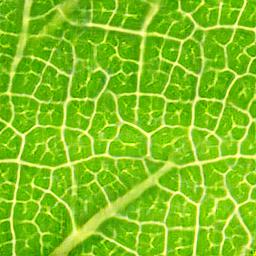} \\
    \end{center}
\caption{{\bfseries Large texture synthesis.} (Top) We collect a co-occurrence tensor from the input image (left) and use it to synthesize an image of a different size (right). (Bottom) 
Synthesizing multiple texture crops and tiling them together controls the local properties of texture at the cost of noticeable seams (left example in each pair).  The generator, however, is able to create a single large texture image that respects local statistics while avoiding the noticeable seams (right example in each pair).}
\label{fig:stitching_interpolations}
\end{figure*}

In Figure \ref{fig:stitching_interpolations}, we compare our generated images to ones obtained by simply tiling multiple synthesized texture crops. As the figure illustrates, this patch-based solution creates noticeable seams. Feeding the same co-occurrences simultaneously to our fully convolutional architecture yields a coherent result, with smooth transitions between the local properties of the synthesized crops.

\begin{figure*}
\begin{center}
\includegraphics[width=1.0\textwidth]{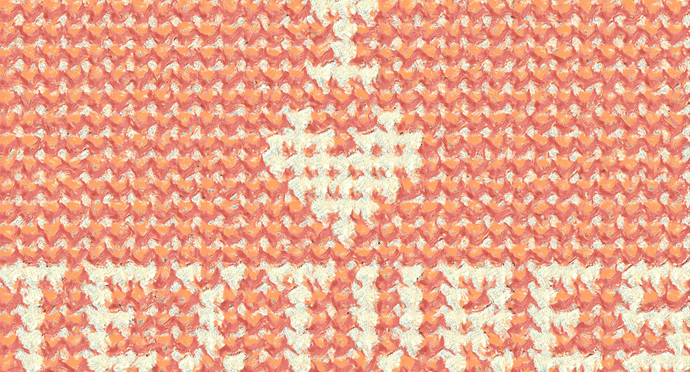}
\\ 
\vspace{0.5cm}
\includegraphics[width=1.0\textwidth]{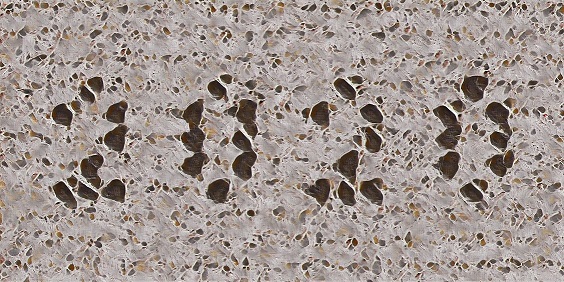}
\caption{{\bfseries Generating large textures with control.} Textures of size $3968 \times 2176$ pixels (top) and $2816 \times 1408$ pixels (bottom), produced by our method. We can control the local appearance of the texture, by constructing the co-occurrence condition to the generator network. For these examples, we used co-occurrences from \textit{only} two $128 \times 128$ crops of the original textures!}
\label{fig:large_supp_1}
\end{center}
\end{figure*}
\section{Controllable Textures}
We show additional results of large controllable textures in Figure \ref{fig:large_supp_1}.

\end{document}